\documentclass[journal]{IEEEtran}

\ifCLASSINFOpdf
\else
\fi
\usepackage{mathtools}
\usepackage{multirow}
\usepackage[font=footnotesize]{subcaption}
\usepackage[font=footnotesize]{caption}
\bstctlcite{IEEEexample:BSTcontrol}
\usepackage{amsthm}
\usepackage{graphicx}
\usepackage{color}
\usepackage{epstopdf}
\usepackage{amsfonts}
\usepackage{cite}
\usepackage{hyperref}
\usepackage{amssymb}
\usepackage{amsmath}
\usepackage{algpseudocode,algorithm}
\usepackage{setspace}
\graphicspath{{Figures/}}
\newtheorem{lemma}{Lemma}

\newtheorem{theorem}{Theorem}

\IEEEaftertitletext{\vspace{-2\baselineskip}}

\usepackage{tikz}
\definecolor{lime}{HTML}{A6CE39}
\DeclareRobustCommand{\orcidicon}{%
    \begin{tikzpicture}
    \draw[lime, fill=lime] (0,0) 
    circle [radius=0.16] 
    node[white] {{\fontfamily{qag}\selectfont \tiny ID}};    \draw[white, fill=white] (-0.0625,0.095) 
    circle [radius=0.007];    \end{tikzpicture}
    \hspace{-2mm}}
\foreach \x in {A, ..., Z}{%
    \expandafter\xdef\csname orcid\x\endcsname{\noexpand\href{https://orcid.org/\csname orcidauthor\x\endcsname}{\noexpand\orcidicon}}
    }

\begin{document}
\title{Federated Learning with Erroneous Communication Links}

\author{Mahyar~Shirvanimoghaddam\orcidA{},~\IEEEmembership{Senior~Member,~IEEE}, Ayoob~Salari\orcidB{},,~\IEEEmembership{Graduate~Student~Member,~IEEE}, Yifeng~Gao, Aradhika~Guha
\thanks{The authors are with the School of Electrical and Information Engineering, The University of Sydney, NSW 2006, Australia. E-mails: mahyar.shirvanimoghaddam@sydney.edu.au, ayoob.salari@sydney.edu.au, aguh5894@uni.sydney.edu.au, ygao6592@uni.sydney.edu.au.}}


\maketitle

\begin{abstract}
In this paper, we consider the federated learning (FL) problem in the presence of communication errors. We model the link between the devices and the central node (CN) by a packet erasure channel, where the local parameters from devices are either erased or received correctly by CN with probability $\epsilon$ and $1-\epsilon$, respectively. We proved that the FL algorithm in the presence of communication errors, where the CN uses the past local update if the fresh one is not received from a device, converges to the same global parameter as that the FL algorithm converges to without any communication error. We provide several simulation results to validate our theoretical analysis. We also show that when the dataset is uniformly distributed among devices, the FL algorithm that only uses fresh updates and discards missing updates might converge faster than the FL algorithm that uses past local updates.
\end{abstract}



\section{Introduction}

\IEEEPARstart{I}{nternet} of things (IoT) applications and services have become popular in recent years due to major technological advancements in sensing, communications, and computation \cite{zhang2021survey}. In many IoT applications, devices operate with extremely low power due to the size, cost, and technological limitations. Traditionally, the devices upload their data to a central node (CN), where all data gathered will be analyzed together. However, such an approach is not feasible in many IoT settings, due to privacy issues, limited power, and usually insufficient communication bandwidth \cite{imteaj2021survey}. An alternative solution is to use federated learning (FL) \cite{wahab2021federated}, where each user is responsible for computing the updates to the current global model based on its own local training data and transmitting them to the CN. Using the updates from the devices, CN improves the global model and redistributes the updated global model to the users. This process is repeated until the convergence is achieved \cite{jin2022communication}.


In FL, communication between devices and CN can be significantly slower than local computing \cite{9272666,zhao2021federated}. Since devices need to communicate with CN repeatedly until the convergence is acquired, there exists a trade-off between local computational power and communication overhead. As we increase the number of local iterations, the required communication between users and CN reduces, and vice versa. The effect of large communication overhead on the scalability of FL and trade-off between local updates and global aggregation was investigated in \cite{wang2018giant,8664630}. Another challenge of FL is the system heterogeneity in which different devices have different computational capabilities, power levels, and storage capacities. In \cite{9187874}, wight-based federated averaging was proposed to tackle this issue.

The majority of works on FL examined a simplified model for the communication channel, where transmissions are error-free but rate-limited \cite{amiri2021convergence,ren2020accelerating,yang2019scheduling}. However, since many IoT devices are operating with low power, have limited computational capabilities, and send short packets, the wireless channel between IoT devices and CN is usually erroneous. Recently, a few studies considered FL over a fading wireless channel \cite{zhang2020federated}. In \cite{amiri2020federated}, to enhance the learning accuracy over a fading channel, a new scheme was proposed, in which at each iteration, based on the channel state, only one device is selected for transmission using a capacity-achieving channel code. However, there is still a lack of full understanding of the effect of communication error on the accuracy and convergence of FL approaches. This is particularly important for IoT applications, since the channel between devices and CN is usually weak.  Due to the large number of devices, the local updates from some of the devices may not reach the CN, which deteriorates the FL performance. In this paper, we shed some light on the effect of communication errors on the convergence and accuracy of FL algorithms. 

We consider a distributed learning problem, where the links between the devices and the CN is modeled by packet erasure channels. In particular, we assume the local updates sent by devices face communication errors; that is, the update is erased or received successfully with a certain probability. We consider two scenarios to calculate the global parameter, where the CN only uses the received fresh local updates and discard missing updates or reuse past updates for devices with missing updates. We further analyse the convergence of these approaches, and prove that by using old local updates in case of errors, the FL algorithm employing gradient descent (GD) converges, and the global parameter will converge to the optimal global parameter. This means that the CN does not necessarily needs fresh updates in every communication round of FL to calculate the global parameter. Instead it can reuse past updates in case of error and continue the FL without jeopardizing the global accuracy. We provide simulation results to verify our analysis and further discuss the convergence behaviour for various datasets with different statistical properties.

The rest of the paper is organized as follows. In Section II, we explain the system model and describe different FL approaches in the presence of communication errors. In Section III, we analyze the convergence and accuracy of the FL approaches in the presence of communication error. Simulation results are provided in Section IV. Finally, Section V concludes the paper.

\section{System Model}
We consider a general federated learning problem, where $N$ device with local datasets, $\mathcal{D}_1$, $\mathcal{D}_2$, $\cdots$, $\mathcal{D}_N$, communicate with the central node (CN). The channel between the $i^{th}$ device and CN is modeled by a packet erasure channel with erasure probability $\epsilon_i$; that is a packet sent from the device to the CN is erased with probability $\epsilon_i$ and received correctly with probability $1-\epsilon_i$. The erasure events for all channels are independent of each other. This model is particularly important for massive IoT scenarios, where many IoT devices communicate with a CN using short packets and due to limited power at devices, the communication channel is mostly erroneous. In particular, let us assume that the local updates from each device is a packet of length $k$ bits, which is encoded using a channel code of rate $R=k/n$, where $n$ is the codeword length. By using the normal approximation bound \cite{Polyanskiy2010}, the packet error rate at the receiver, when the signal-to-noise ratio is $\gamma$, is given by:
\begin{align}
    \epsilon\approx Q\left(\frac{n\log_2(1+\gamma)-k+\log_2(n)}{\sqrt{nV(\gamma)}}\right),
\end{align}
where $Q(.)$ is the standard $Q$-function and $V(\gamma)=\left(1-(1+\gamma)^{-2}\right)\log_2^2(e)$ is the channel dispersion \cite{Polyanskiy2010}. The short packet communication can then be modeled by the packet erasure channel with erasure probability $\epsilon$, when $k$, $n$, and $\gamma$ is known. Using this simplified model, the channel quality can be characterized by a single parameter $\epsilon$. 

We also assume that the downlink channel, i.e., the channel from the CN to devices, is error-free since the CN can transmit with high power; therefore, the error in the downlink channel can be appropriately mitigated.

\vspace{-4ex}
\subsection{Federated Learning in the error-free scenario}
In FL, each device $i$ calculates the local update and sends the parameters to the CN. Then, CN aggregates all the parameters received from all nodes and calculates the general parameters. In particular, let $w^{(t)}$ denote the local parameter calculated at device $i$ at time instant $t$. By using the gradient descent (GD) method with learning rate $\eta$ and local loss function $F_i:\mathbb{R}^d\to\mathbb{R}$, $w^{(t)}_i$ can be calculated as follows:
\begin{equation}
w^{(t)}_i = w^{(t-1)} - \eta \nabla F_i(w^{(t-1)}).
\end{equation}
Upon receiving all local parameters (assuming that $\epsilon_i=0$ for all devices) from all devices, the CN calculates the global parameter as follows:
\begin{equation}
w^{(t)} = \frac{1}{D}\sum_{i=1}^{N}D_iw^{(t)}_i,
\end{equation}
where $D_i=|\mathcal{D}_i|$ is the size of dataset $\mathcal{D}_i$ and $D=\sum_{i=1}^ND_i$.

\vspace{-3ex}
\subsection{Federated Learning in the presence of communication error}
Here, we consider that the channel between the nodes and the CN is erroneous and that the local updates calculated from some of the nodes may not reach the CN. 
\subsubsection{FL with erroneous communication and no memory at CN}
In this scenario, the CN calculates the global parameter using the received local updates only. The global update, in this case, is calculated as follows:
\begin{align}
w^{(t)} = \frac{1}{\sum_{i\in\mathcal{S}(t)}D_i}\sum_{i\in\mathcal{S}(t)}D_iw^{(t)}_i,
\label{eq:globalnomemory}
\end{align}
where $\mathcal{S}(t)$ is the set of all nodes that their updates have been successfully received at CN at time instant $t$. 
\subsubsection{FL with erroneous communication and reuse of old local updates}
We consider another scenario, in which the CN uses the previous updates received from the devices in case their new updates are not received. In this case, the global update is calculated as follows:
\begin{align}
w^{(t)} = \frac{1}{D}\left(\sum_{i\in\mathcal{S}(t)} D_iw^{(t)}_i+\sum_{j\in\mathcal{F}(t)}D_jw^{(t-1)}_j\right),
\label{eq:globalreuse}
\end{align}
where $\mathcal{F}(t)$ is the set of all nodes that their updates have not been received at CN at time instant $t$. Here, we assumed that the previous local updates for missing nodes are always available at the CN. This assumption is valid as long as the erasure probability is low. 

\begin{figure}[!t]
     \centering
     \begin{subfigure}[t]{0.69\columnwidth}
         \centering
         \includegraphics[width=0.95\columnwidth]{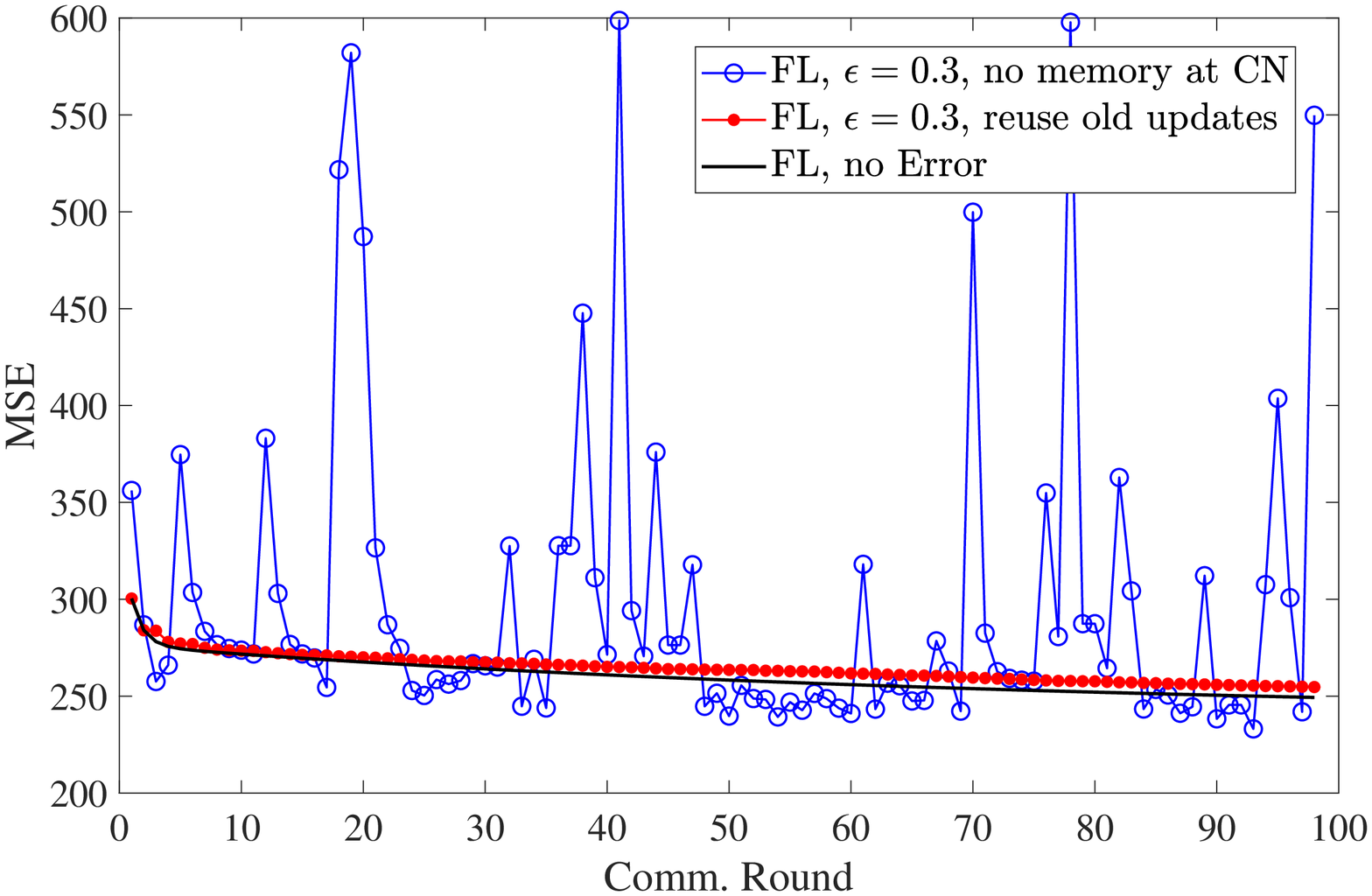}
         \caption{MSE vs. Comm. round.}
         \label{fig:comp1mse}
     \end{subfigure}
     \hfill
     \begin{subfigure}[t]{0.29\columnwidth}
         \centering
         \includegraphics[width=0.95\columnwidth]{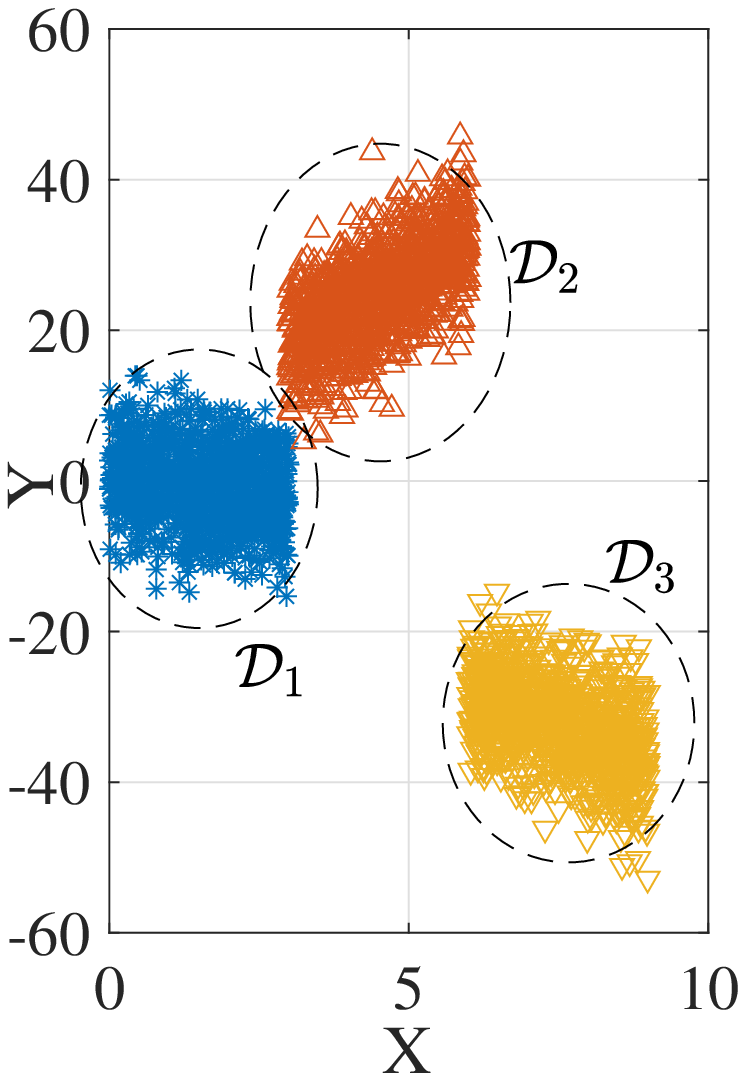}
         \caption{Datasets.}
         \label{fig:dataset1}
     \end{subfigure}
        \caption{A comparison between error-free and erroneous FL, when the number of devices is $N=3$, $|D_i|=1000$, and GD with learning rate $\eta=0.005$ and maximum 10 iterations at devices is used.}
        \label{fig:comp1}
\end{figure}

Fig. \ref{fig:comp1} shows a comparison between the FL schemes with and without communication errors. As can be seen in this figure, when CN does not have a memory, the overall mean squared error (MSE) fluctuates and the FL does not converge.  However, when CN can store previous updates from the devices, it can reuse them if the fresh updates are missing (see \eqref{eq:globalreuse}). In this case, the FL algorithm converges to the same MSE as that for FL without any communication error. 

\vspace{-2ex}
\section{Performance Analysis of the FL Algorithm in the Presence of Communication Error}
For the simplicity of the analysis, we assume that the size of all local datasets are the same, i.e., $D_i=D/N$, for $i=1,\cdots,N$. When the CN does not have memory, all updates which have not been received will be discarded from the global aggregation step. Let us, for a moment, assume that the number of local iterations at the nodes is sufficiently large. That is, when each node performs a local update, the GD at that device converges. In this case, we define $w^{(t)}_i=w_i$ for $i=1,\cdots,N$. By using \eqref{eq:globalnomemory}, the global update at time instant $t$ is given by:
\begin{align}
w^{(t)} = \frac{1}{|\mathcal{S}(t)|}\sum_{i\in\mathcal{S}(t)}w_i.
\label{eq:convnomemory}
\end{align}
Since the link between the $i^{th}$ device and CN is an erasure channel with erasure probability $\epsilon_i$, the probability mass function $(\mathrm{pmf})$ of $w^{(t)}$ can be calculated as follows:
\begin{align}
\mathrm{Prob}\left\{w^{(t)}=\frac{\sum_{i=1}^NI_iw_i}{\sum_{i=1}^NI_i}\right\} =\prod_{i=1}^N\epsilon_i^{1-I_i}(1-\epsilon_i)^{I_i}, 
\label{eq:pmfnomemory}
\end{align}
where $I_i\in\{0,1\}$ denote the erasure event for the link between the $i^{th}$ device and CN, i.e., $\mathrm{Prob}\{I_i=1\}=1-\epsilon_i$ and $\mathrm{Prob}\{I_i=0\}=\epsilon_i$. From \eqref{eq:pmfnomemory} it can be easily observed that the global parameter always fluctuates due to the random nature of the erasure events. This can be clearly seen in Fig. \ref{fig:comp1mse}. 

\vspace{-2ex}
\subsection{Performance Analysis of the FL algorithm  with erroneous communication and reuse of old local updates}
Here, we assume that $F_i(.)$, for all $i\in\{1,\cdots,N\}$, is convex and $L$-smooth. For function $f:\mathbb{R}^d\to\mathbb{R}$ that is convex and $L$-smooth, the following relations hold, $\forall x,y\in\mathbb{R}^{d}$ \cite{MAL-050}:
\begin{align}
    \label{convexity} 
    f(y)&\le f(x)+\nabla f(x)(y-x)'+\frac{L}{2}||y-x||_2^2,\\
    \label{smoothness}
    f(x^*)&-f(x)\le -\frac{1}{2L}||\nabla f(x)||_2^2, ~x^*=\arg\min_x f(x),\\
    \label{smoothconvex}
    ||\nabla f(&x)- \nabla f(y)||_2\le L||x-y||_2,\\
    \left(\nabla f(\right.&x)-\left.\nabla f(y)\right)(x-y)'\ge\frac{1}{L}||\nabla f(x)-\nabla f(y)||_2^2,
    \label{smoothc2}
\end{align}
where $(.)'$ denote the matrix transpose operand.  
\begin{lemma}
Let $F_i:\mathbb{R}^d\to \mathbb{R}$, for all $i\in\{1,\cdots,N\}$, is convex and $L$-smooth. $F_{\mathcal{G}}(x)=(1/N)\sum_{i\in\mathcal{G}}F_i(x)$ is also convex and $\frac{|\mathcal{G}|L}{N}$-smooth.
\end{lemma}
\begin{IEEEproof}
This can be easily proved from the sub-additivity of the norm and additivity of the gradient \cite{MAL-050}.
\end{IEEEproof}
From Lemma 1, it can be easily proved that $F(x)=(1/N)\sum_{i=1}^N F_i(x)$ is also convex and $L$-smooth. In the following theorem, we show that the FL algorithm in the presence of communication error, where the CN uses the past local updates in case of communication error, converges to the global minima of the global loss function and accordingly the optimal global parameter.
\begin{theorem}
Let us consider a FL problem with $N$ devices, where the channel from each device to the CN is modeled by an erasure channel with erasure probability $\epsilon$. We assume that the local loss function $F_i(x)$ at device $i$ is convex and $L$-smooth. We further assume that $||\nabla F(x)- \nabla F(y)||_2\ge\mu||x-y||_2$, for all $x,y\in \mathbb{R}^d$, where $F(x)=\frac{1}{N}\sum_{i=1}^NF_i(x)$. Let $\delta_t=||w^{(t)}-w^*||_2^2$, where $w^*=\arg \min_w F(w)$, and $\bar{\delta}_{t+1}=\frac{1}{t+1}\sum_{i=0}^{t}\delta_i$. For the FL algorithm \eqref{eq:globalreuse}, when $\epsilon\le\frac{\mu}{2L}$ and $\eta=\frac{1}{L}$, $\bar{\delta}_{k}$ is upper bounded by:
\begin{align}
    \bar{\delta}_{t}\le\frac{F(w^{(0)})-F(w^*)}{t\beta^2},~~\text{for}~t>0,
\end{align}
where $\beta^2=\frac{\mu^2}{2L}-2L\epsilon^2$.
\end{theorem}
\begin{IEEEproof}
The proof is provided in Appendix A.
\end{IEEEproof}
Theorem 1 states that the gap to the global minima ($w^*$) decreases with the iteration number $t$ and converges to zero, when $t$ is arbitrary large, i.e., $\lim_{t\to\infty} \bar{\delta}_{t}=0$. it is also easy to show that $\lim_{t\to\infty} \delta_{t}=0$, since otherwise, if $\delta_{t}\ge\epsilon$, where $\epsilon>0$,  $\bar{\delta}_{t}$ will be always bounded above $\epsilon$.  It is important to note that Theorem 1 does not state that $\delta_t$ is a decreasing function of $t$. Instead it shows that in the limit that $t$ is sufficiently large, the global parameter, $w^{(t)}$, converges to the optimal global parameter $w^*$, even in the presence of communication error.

\section{Results and Discussion}
\begin{figure}[t]
         \centering
         \includegraphics[width=0.9\columnwidth]{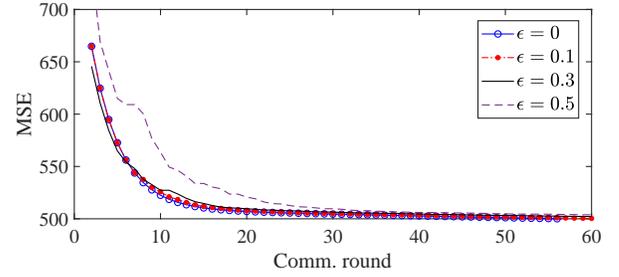}
         \caption{A comparison between error-free and erroneous FL, when the number of devices is $N=3$, $|D_i|=1000$, and GD with learning rate $\eta=0.005$ and maximum 1 iterations at devices is used.}
         \label{fig:comp3mse}
         \vspace{-1ex}
\end{figure}

In this section, we focus on the FL algorithm, where the CN uses old local updates received from devices when their fresh updates are not available due to communication error. Fig. \ref{fig:comp3mse} shows the overall MSE of the FL algorithm at various erasure probabilities, when $N=3$. We use the same dataset as in Fig. \ref{fig:dataset1}. As can be seen in Fig. \ref{fig:comp3mse}, when the erasure rate is small, i.e., $\epsilon=0.1$, the performance of the FL with reuse of old updates closely approaches that of the FL without communication errors. However, when the erasure probability increases, the MSE in the early iterations has a significant gap to the ideal FL case with no error. In all scenarios, even when the erasure rate is large, i.e., $\epsilon=0.5$, the FL algorithm with reuse of old updates in the presence of error converges to that of the ideal FL without error.

\begin{figure}[!t]
     \centering
     \begin{subfigure}[t]{0.69\columnwidth}
         \centering
         \includegraphics[width=0.95\columnwidth]{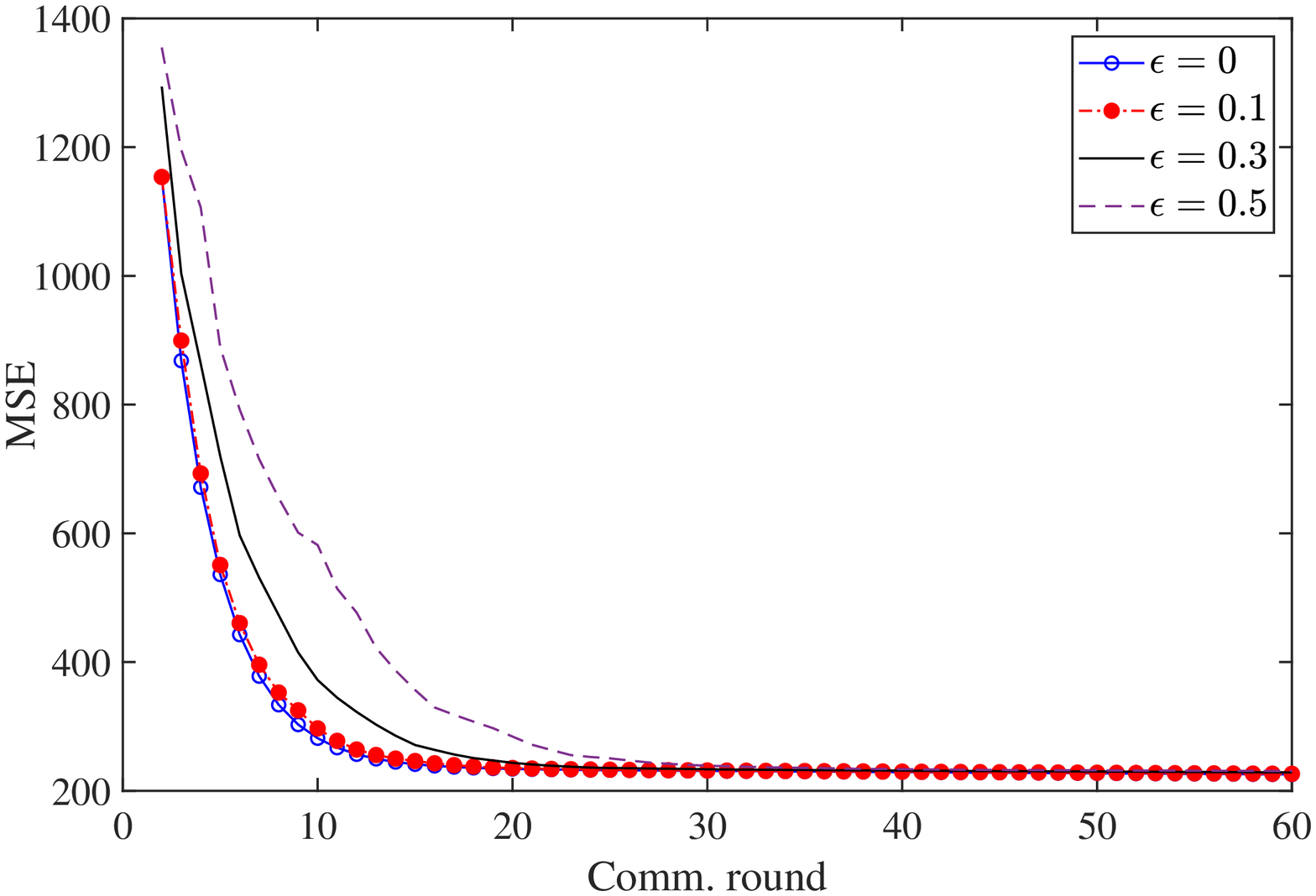}
         \caption{MSE vs. Comm. round.}
         \label{fig:comp10mse}
     \end{subfigure}
     \hfill
     \begin{subfigure}[t]{0.29\columnwidth}
         \centering
         \includegraphics[width=0.95\columnwidth]{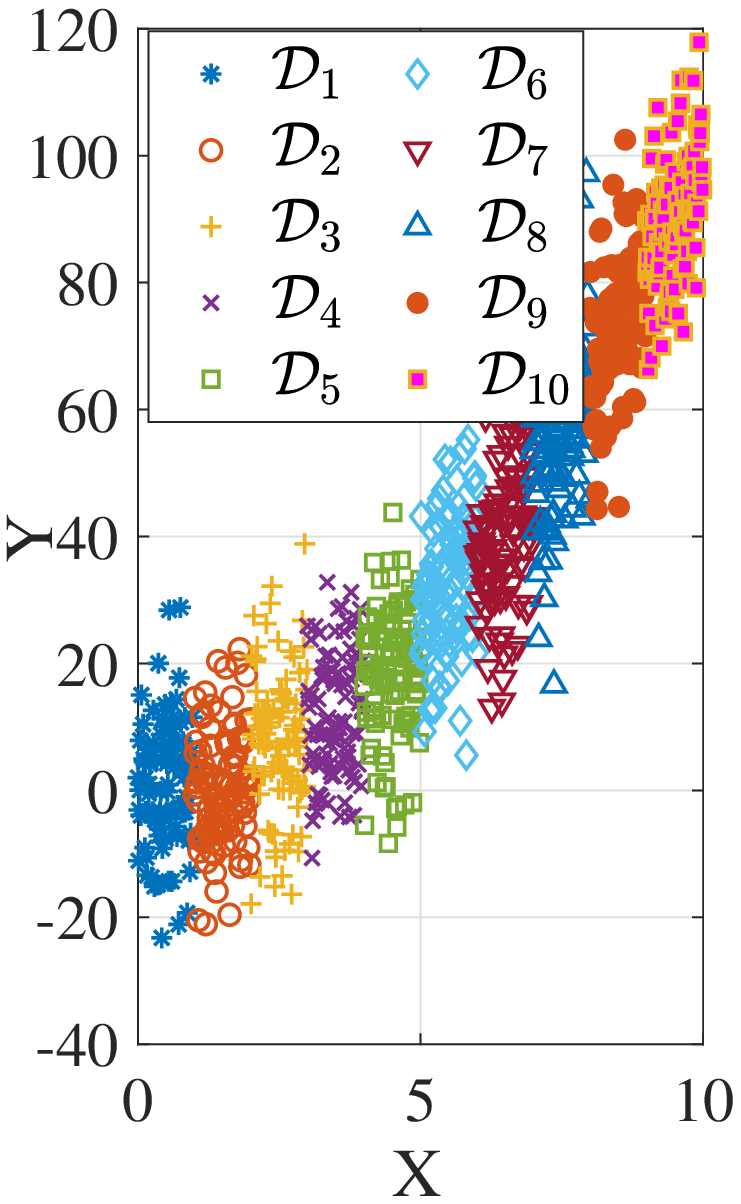}
         \caption{Datasets.}
         \label{fig:dataset10}
     \end{subfigure}
     \vspace{-1ex}
        \caption{A comparison between error-free and erroneous FL, when the number of devices is $N=10$, $|D_i|=100$, and GD with learning rate $\eta=0.005$ and maximum 1 iterations at devices is used.}
        \label{fig:comp10}
        \vspace{-0.5em}
\end{figure}

Fig. \ref{fig:comp10mse} shows the performance of FL algorithm with reuse of old updates at various erasure probabilities, when the number of devices is $N=10$. The datasets are shown in Fig. \ref{fig:dataset10}. The data is created by using a non-linear model $y=x^2+z$, where $z\sim\mathcal{N}(0,\sigma^2)$, is additive white Gaussian noise. As can be seen in  Fig. \ref{fig:dataset10}, the linear regression curve for each local dataset has a different slope; therefore, losing any dataset due to error will result in a significant change in the global parameter. This model is of particular importance for IoT scenarios, where devices are distributed in a field at various locations; therefore, their measurements are location/time dependent and accordingly their datasets are non-iid. As can be seen in Fig. \ref{fig:comp10mse}, when the CN reuses past local updates in case of error, the MSE performance of the FL algorithm converges to that of the ideal FL algorithm without error. 

\begin{figure}[!t]
     \centering
     \begin{subfigure}[t]{0.69\columnwidth}
         \centering
         \includegraphics[width=0.95\columnwidth]{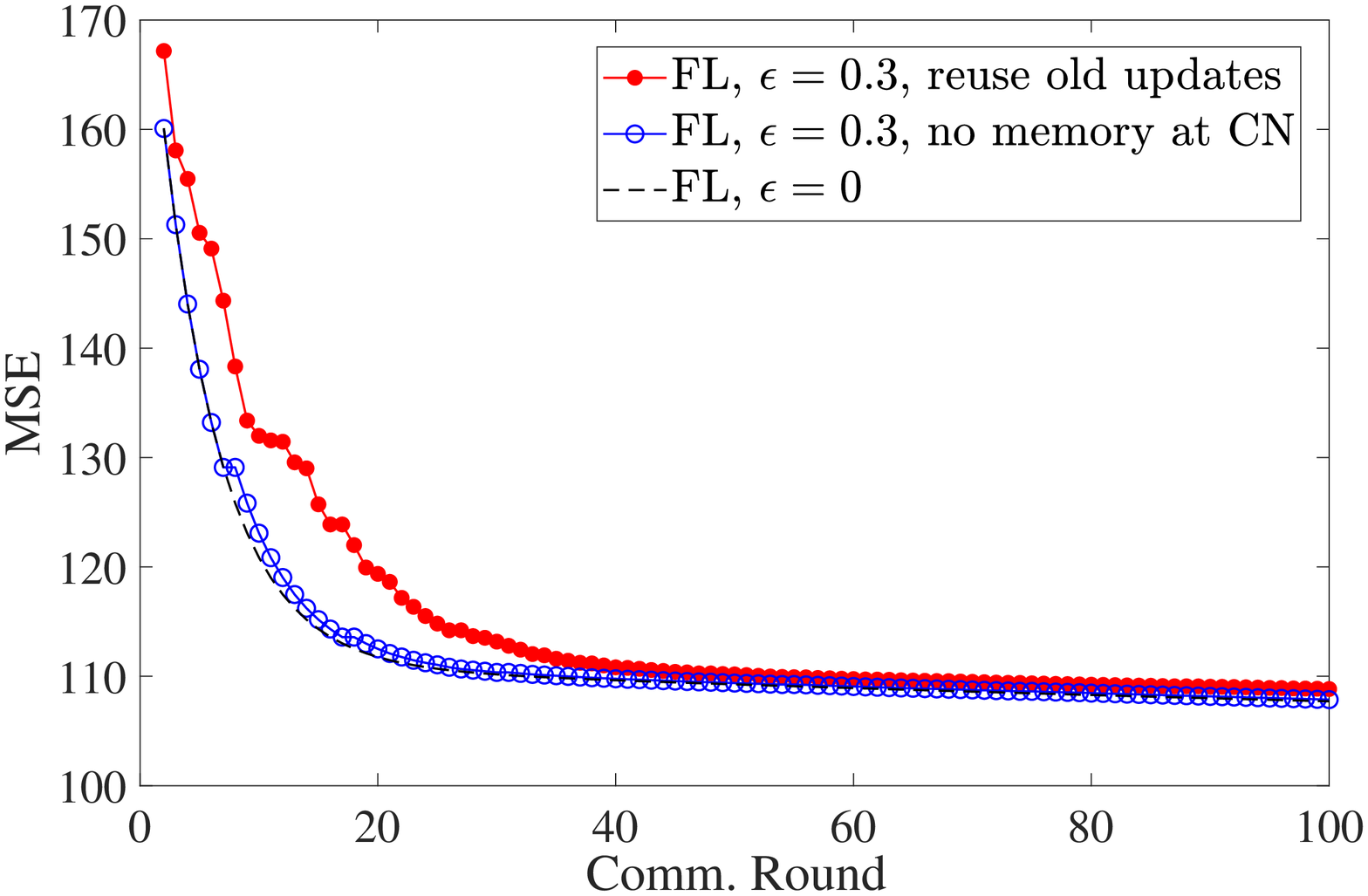}
         \caption{MSE vs. Comm. round.}
         \label{fig:comp1mseunif}
     \end{subfigure}
     \hfill
     \begin{subfigure}[t]{0.29\columnwidth}
         \centering
         \includegraphics[width=0.95\columnwidth]{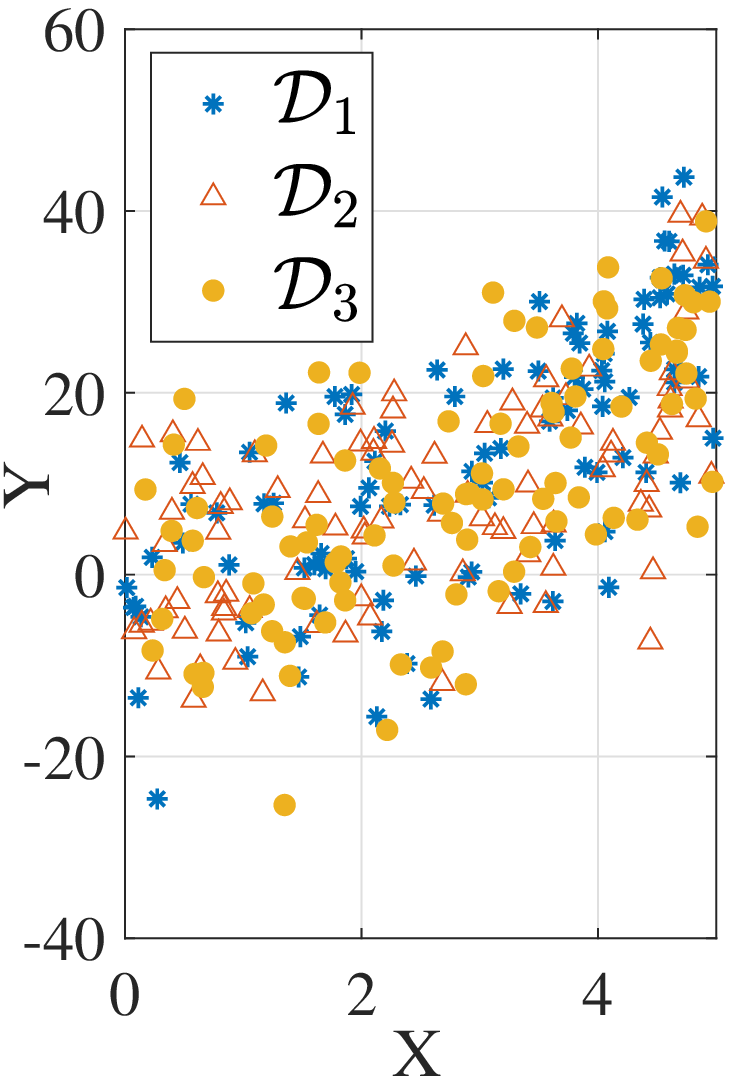}
         \caption{Datasets.}
         \label{fig:datasetunif}
     \end{subfigure}
     \vspace{-1ex}
        \caption{A comparison between error-free and erroneous FL, when the number of devices is $N=3$, $|D_i|=100$, and GD with learning rate $\eta=0.01$ and maximum 1 iterations at devices is used.}
        \label{fig:compunif}
        \vspace{-0.5em}
\end{figure}

It is important to note that when the dataset is uniformly distributed among devices and the local parameters are not significantly different, even without reusing old local updates, the FL algorithm converges. An example is provided in Fig. \ref{fig:compunif}, where the dataset is uniformly distributed between 3 devices. In this case, the local parameter from all devices are relatively close to each other. Therefore, missing some devices' updates does not affect the overall performance. As can be seen in Fig. \ref{fig:comp1mseunif}, the FL algorithm without memory at the CN outperforms the FL algorithm with reusing local updates, when the dataset is uniformly distributed among devices.  This can be explained for an extreme case, when $D$ is relatively large, no error occurs up until the $i$th communication round, and all local updates are the same in each round. In this case, by using \eqref{eq:globalnomemory}, we will have $w^{(t)}=w^{(t)}_i=w^{(t-1)}-\eta\nabla F_i(w^{(t-1)})$. Since the datasets are uniformly distributed, $w^{(t)}=w^{(t-1)}-\eta\nabla F(w^{(t-1)})$, even if an error occurs. However, if in the case of error we reuse old updates, by using \eqref{eq:globalreuse} the global parameter will be $w^{(t)}=w^{(t)}_i-\epsilon\left(w^{(t)}_i-w^{(t-1)}_i\right)$. This is equivalent to $w^{(t)}=w^{(t-1)}-\eta\nabla F(w^{(t-1)})-\epsilon\left(w^{(t)}_i-w^{(t-1)}_i\right)$, which means that the global parameter will have a gap, proportional to the erasure rate, to that of the error-free case. Therefore, for uniformly distributed datasets, by reusing old updates in case of communication error, the FL algorithm takes longer to converge.

\begin{figure}[t]
         \centering
         \includegraphics[width=0.9\columnwidth]{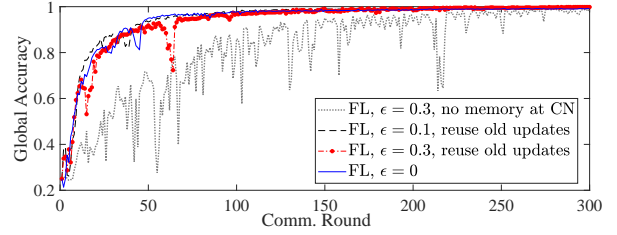}
         \vspace{-1ex}
         \caption{Test accuracy of FL using the MNIST dataset and $N=4$ devices.}
         \label{fig:cifar10}
         \vspace{-1ex}
\end{figure}

We further consider an image classification task using a real dataset from MNIST \cite{mcmahan2017communication}, which consists of 4000 handwritten images of the numbers $0$ to $3$. The example runs in parallel using 4 workers (i.e., devices), each processing images of a single digit. In particular, worker $i$ has 700 handwritten images of number $i-1$ as its training set. The validation and test set at the CN each has 150 images of each number $0$ to $3$. Similar to \cite{mcmahan2017communication}, at each worker, we use a CNN with two $5\times 5$ convolution layers (the first with 32 channels, the second with 64, each followed with $2\times 2$ max pooling), a fully connected layer with 512 units and ReLu activation, and a final softmax output layer. We also use the stochastic GD optimizer with learning rate $0.001$. Fig. \ref{fig:cifar10} shows the accuracy of the FL algorithm after each communication round in the presence of communication error. As can be seen in this figure, when the CN uses the old local updates if the fresh updates are erased due to communication error, the FL algorithm converges to the same level of accuracy as that for the FL algorithm without any communication error. However, when the CN does not store past local updates and only uses fresh updates and discard missing updates in the aggregation stage, the accuracy changes significantly with the communication round.

\section{Conclusions}
In this paper, we studied the federated learning algorithm in the presence of communication errors. We modelled the channel between the devices and the central node (CN) by packet erasure channels. We presented two approaches to deal with error events. That is when an error occurs, and local updates are not received at CN, the CN can either calculate the global parameter by using only the fresh local updates received correctly or reuse the old updates for missing local updates. We proved that the FL algorithm with reusing local updates in case of error converges to the same global parameters as that with FL without any communication error. This means that the CN does not need to wait for the retransmission of missing updates. Instead, it can reuse the past local updates to calculate the global parameters and continue the FL algorithm. This approach converges to the same result as FL achieves with no communication error. This is of practical importance as IoT devices can significantly save energy by relaxing the communication reliability requirements without jeopardizing the overall learning accuracy. We further provided simulation results to verify our theoretical results. We also highlighted that when the dataset is uniformly distributed among devices, the FL algorithm without memory at the CN converges faster than that with reusing local updates.
\appendices
\section{Proof of Theroem 1}
Assuming that $D_i=D/N$, \eqref{eq:globalreuse} can be written as follows:
\begin{align}
w^{(t+1)}&= \frac{1}{N}\sum_{i\in\mathcal{S}(t)}w^{(t)}_i+\frac{1}{N}\sum_{j\in\mathcal{F}(t)}w^{(t-1)}_j\\
\nonumber&=\frac{1}{N}\sum_{i\in\mathcal{S}(t)}\left(w^{(t)}-\eta\nabla F_i(w^{(t)})\right)\\
\nonumber&+\frac{1}{N}\sum_{j\in\mathcal{F}(t)}\left(w^{(t-1)}-\eta\nabla F_j(w^{(t-1)})\right)\\
\nonumber&=w^{(t)}-\eta\nabla F(w^{(t)})+\frac{|\mathcal{F}(t)|}{N}\left(w^{(t-1)}-w^{(t)}\right)\\
&+\eta\nabla F_{\mathcal{F}}(w^{(t)})-\eta\nabla F_{\mathcal{F}}(w^{(t-1)}),
\label{eq:iteraton}
\end{align}
where $F_{\mathcal{F}}(x)=\frac{|\mathcal{F}(x)|}{N}\sum_{j\in\mathcal{F}(t)}F_j(x)$. Since $F(x)$ is convex and $L$-smooth, by using \eqref{convexity}, we have:
\begin{align}
    \nonumber F(&w^{(t+1)})\le F(w^{(t)})+\nabla F(w^{(t)})\left(w^{(t+1)}-w^{(t)}\right)'\\
    \nonumber &+\frac{L}{2}||w^{(t+1)}-w^{(t)}||_2^2\\
    \nonumber &\overset{\eqref{eq:iteraton}}{=}F(w^{(t)})-\eta||\nabla F(w^{(t)})||_2^2\\
    \nonumber&+\frac{|\mathcal{F}(t)|}{N}\nabla F(w^{(t)})(w^{(t-1)}-w^{(t)})'\\
    \nonumber &+\eta\nabla F(w^{(t)})(\nabla F_{\mathcal{F}}(w^{(t)})-\nabla F_{\mathcal{F}}(w^{(t-1)}))'\\
    \nonumber &+\frac{L}{2}\eta^2||\nabla F(w^{(t)})||_2^2+\frac{L|\mathcal{F}(t)|^2}{2N^2}||w^{(t-1)}-w^{(t)}||_2^2\\
    \nonumber &+\frac{L}{2}\eta^2||\nabla F_{\mathcal{F}}(w^{(t)})-\nabla F_{\mathcal{F}}(w^{(t-1)})||_2^2\\
    \nonumber &-\frac{L|\mathcal{F}(t)|}{N}\eta\nabla F(w^{(t)})(w^{(t-1)}-w^{(t)})'\\
    \nonumber &-L\eta^2\nabla F(w^{(t)})(\nabla F_{\mathcal{F}}(w^{(t)})-\nabla F_{\mathcal{F}}(w^{(t-1)}))'\\
    \nonumber &-\frac{\eta L|\mathcal{F}(t)|}{N}(\nabla F_{\mathcal{F}}(w^{(t)})-\nabla F_{\mathcal{F}}(w^{(t-1)}))(w^{(t)}-w^{(t-1)})'
\end{align}
Now, assuming that $\eta=\frac{1}{L}$ and due to the fact that $|\mathcal{F}_F|\approx \epsilon N$, when $N$ is sufficiently large, this can be simplified to:
\begin{align}
    \nonumber &F(w^{(t+1)})\le F(w^{(t)})-\frac{1}{2L}||\nabla F(w^{(t)})||_2^2\\
    \nonumber &+\frac{L\epsilon^2||w^{(t)}-w^{(t-1)}||_2^2}{2}
    +\frac{||\nabla F_{\mathcal{F}}(w^{(t)})-\nabla F_{\mathcal{F}}(w^{(t-1)})||_2^2}{2L}\\
    \nonumber &-\epsilon(\nabla F_{\mathcal{F}}(w^{(t)})-\nabla F_{\mathcal{F}}(w^{(t-1)}))(w^{(t)}-w^{(t-1)})'\\
    \nonumber &\overset{(a)}{\le} F(w^{(t)})-\frac{1}{2L}||\nabla F(w^{(t)})||_2^2+\frac{L\epsilon^2}{2}||w^{(t)}-w^{(t-1)}||_2^2\\
    \nonumber &-\frac{1}{2L}||\nabla F_{\mathcal{F}}(w^{(t)})-\nabla F_{\mathcal{F}}(w^{(t-1)})||_2^2\\
    &\le F(w^{(t)})-\frac{||\nabla F(w^{(t)})||_2^2}{2L} +\frac{L}{2}\epsilon^2||w^{(t)}-w^{(t-1)}||_2^2,
\end{align}
where step $(a)$ follows from \eqref{smoothc2} and Lemma 1, which indicates that $F_{\mathcal{F}}(.)$ is convex and $Le$-smooth. Since we assumed that $||\nabla F(x)- \nabla F(y)||_2\ge\mu||x-y||_2$, for all $x,y\in\mathbb{R}^d$, we have:
\begin{align}
    \nonumber F(w^{(t+1)})&\le F(w^{(t)})-\frac{\mu^2}{2L}||w^{(t)}-w^*||_2^2\\
    &+\frac{L}{2}\epsilon^2||w^{(t-1)}-w^{(t)}||_2^2.
    \label{2nditeration}
\end{align}
It is easy to show that $||w^{(t-1)}-w^{(t)}||_2^2\le 2\left(\delta_t+\delta_{t-1}\right)$. We can further simplify \eqref{2nditeration} as follows:
\begin{align}
F(w^{(t+1)})\le F(w^{(t)})+\left(L\epsilon^2-\frac{\mu^2}{2L}\right)\delta_t+L\epsilon^2\delta_{t-1}.
\end{align}
Summing up both sides over $t=1,\cdots, k$, and using telescopic cancellation, we have:
\begin{align}
\nonumber F(w^{(k+1)})&\le F(w^{(0)})+(2L\epsilon^2-\frac{\mu^2}{2L})\sum_{i=1}^{k-1}\delta_i\\
&+(L\epsilon^2-\frac{\mu^2}{2L})(\delta_k+\delta_0),
\label{eqtelescope}
\end{align}
where we assumed that the first global update ($t=1$), is calculated without any communications error. That is $F(w^{(1)})\le F(w^0)-\frac{\mu}{2L}\delta_0$. Assuming that $\epsilon\le\frac{\mu}{2L}$, we have $\frac{\mu^2}{2L}-2L\epsilon^2<\frac{\mu^2}{2L}-L\epsilon^2$. Therefore, \eqref{eqtelescope} is simplified to:
\begin{align}
    F(w^{(k+1)})&\le F(w^{(0)})-\beta^2(k+1)\bar{\delta}_{k+1},
\end{align}
where $\beta^2=\frac{\mu^2}{2L}-2L\epsilon^2$. By rearranging the above inequality, we have:
\begin{align}
    \bar{\delta}_{k+1}\le\frac{F(w^{(0)})-F(w^{(k+1)})}{(k+1)\beta^2}.
\end{align}
Since $F(w^*)\le F(w^{(k+1)})$, We will have:
\begin{align}
    \bar{\delta}_{k+1}\le\frac{F(w^{(0)})-F(w^*)}{(k+1)\beta^2}.
\end{align}

\bibliographystyle{IEEEtran}
\footnotesize
\bibliography{IEEEabrv,ref}

\end{document}